%% file: main.tex
\documentclass[hyphens]{article}
\usepackage[final]{neurips_2025} 

\usepackage[utf8]{inputenc}
\usepackage[T1]{fontenc}
\usepackage{hyperref}
\hypersetup{
 colorlinks=true,
 linkcolor=red,
 citecolor=cyan,
 filecolor=magenta,
 urlcolor=magenta,
}
\usepackage{url}
\usepackage{xurl}
\usepackage{booktabs}
\usepackage{amsfonts}
\usepackage{nicefrac}
\usepackage{microtype}
\usepackage[table,dvipsnames]{xcolor}
\usepackage{amsmath}
\usepackage{cleveref}
\usepackage{xspace}
\usepackage{enumitem}
\usepackage{textcomp}
\usepackage{stfloats}
\usepackage{verbatim}
\usepackage{wrapfig}
\usepackage{graphicx}
\usepackage{float}
\usepackage{amssymb}
\usepackage[numbers,sort&compress]{natbib}
\usepackage{tikz}
\usepackage{algorithm}
\usepackage{algpseudocode}
\usepackage{makecell}
\usepackage{multicol,multirow}
\usepackage{threeparttable}
\usepackage{tablefootnote}
\usepackage{pgfplots}
\pgfplotsset{compat=1.18}
\usepackage[labelfont=bf]{caption}
\usepackage{booktabs}
\usepackage{graphicx}
\AtBeginEnvironment{thebibliography}{\raggedright\sloppy}
\hfuzz=\maxdimen
\vfuzz=\maxdimen

\usepackage{fancyhdr}
\renewcommand{\headwidth}{\textwidth}

\renewcommand{\headrulewidth}{0.5pt}
\renewcommand{\headrule}{\vspace{2pt}\hbox to\headwidth{\color{black}\leaders\hrule height \headrulewidth\hfill}}
\pgfplotsset{compat=1.18}

\title{TuringViT: Making SOTA Vision Transformers Accessible to All}

\author{
\colorbox{white}{\includegraphics[height=0.8em]{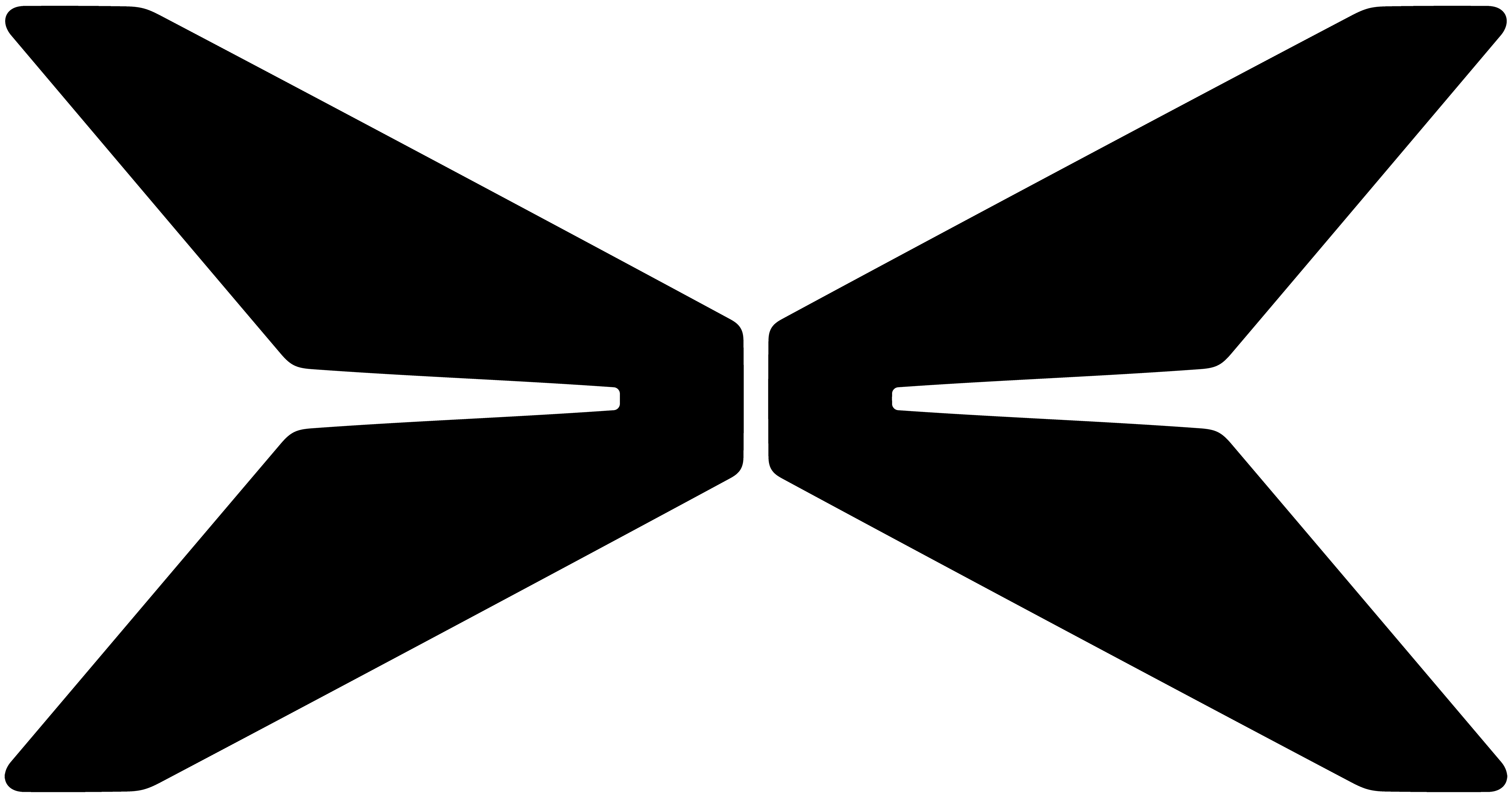}} Foundation Model Team, Xpeng Inc.\\
\href{https://turingvit.github.io/}{\texttt{https://turingvit.github.io}}
}

\vspace{-0.5em} 
\begin{document}
\maketitle
\input{sec/0_abstract}
\input{sec/1_intro}

\input{sec/3_methodology}

\input{sec/4_experiments}
\input{sec/5_application}
\input{sec/6_conclusion}
\input{sec/7_contributors}
{
    \begingroup
    \sloppy
    \small
    \bibliographystyle{turing_unsrtnat}
    \bibliography{main}
    \endgroup
}

\end{document}

%% file: sec/0_abstract.tex
\vspace{-0.8em}
\begin{abstract}
\vspace{-0.8em}

Modern VLMs and VLA systems commonly adopt off-the-shelf ViTs such as SigLIP2 as visual encoders, but diverse downstream requirements in latency, temporal modeling, and VLM integration often call for customized SOTA-level ViTs. Training such encoders remains beyond the reach of much of the community, as it requires massive image-text data, while standard softmax attention makes high-resolution or dynamic-resolution pretraining prohibitively costly and often forces low-resolution pretraining followed by post-hoc adaptation.
\textbf{TuringViT} addresses these challenges with three key designs: \textbf{Turing Linear Attention (TLA) } for efficient sequence modeling, \textbf{VISTA-Curation} to construct supervision-rich image-video training data, and native dynamic-resolution pretraining that supports flexible inputs from the start and transfers seamlessly to downstream VLMs. As a result, TuringViT outperforms leading open-source ViT baselines with only 10\% of the data, achieves stronger downstream VLM performance, and delivers substantially better latency scaling on high-resolution inputs. 
Our scaling-law analysis further shows that TuringViT continues to improve predictably with curated data scale, far from saturation. Its fast adaptation, hardware-friendly design, and efficient deployment have made it a unified visual foundation across XPeng’s AI systems. More broadly, TuringViT provides a reproducible pipeline that dramatically lowers the cost for the community to train, customize, and deploy SOTA-level ViTs, moving toward making such Vision Transformers accessible to all.

\vspace{-1.2em}
\end{abstract}

\begin{center}
    \vspace{-0.6em}
    \includegraphics[width=0.72\textwidth]{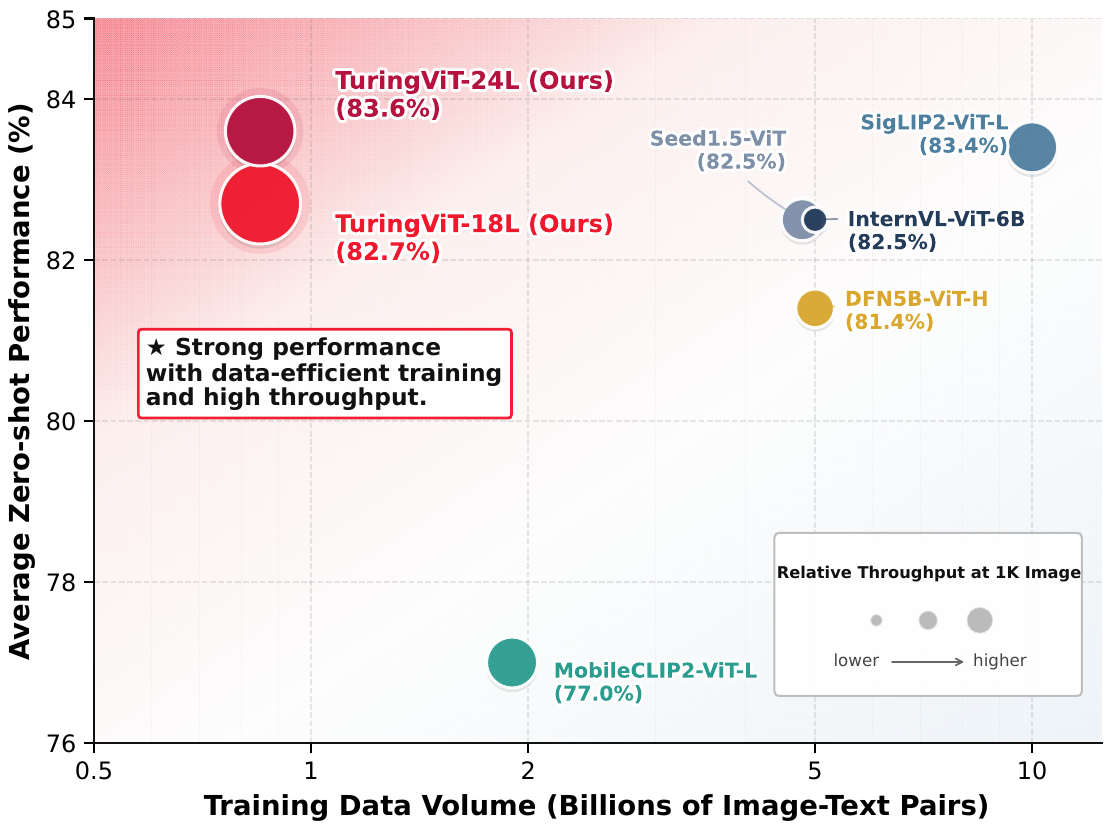}
    \captionof{figure}{\textbf{Zero-shot performance, training data scale, and 1K-resolution throughput of vision transformer encoders.} TuringViT achieves stronger accuracy--efficiency trade-offs than leading open-source baselines with only 10\% of the training data.}
    \label{fig:vit_compare}
    \vspace{-0.8em}
\end{center}

%% file: sec/1_intro.tex
\section{Introduction}
\label{sec:intro}

In modern VLMs~\cite{bai2023qwenvlversatilevisionlanguagemodel,internvl,guo2025seed15vltechnicalreport} and VLA~\cite{intelligence2026pi07steerablegeneralistrobotic, yu2026dm0embodiednativevisionlanguageactionmodel} systems, the visual encoder has become the primary interface through which language models perform increasingly complex multimodal perception, reasoning, and action. A common practice is to adopt powerful off-the-shelf encoders pretrained on large-scale image-text data~\cite{clip,vasu2024mobileclipfastimagetextmodels}. Recent models such as SigLIP2~\cite{tschannen2025siglip2multilingualvisionlanguage} have shown strong general-purpose performance, but simply reusing an existing encoder is not always sufficient. Different downstream settings may impose distinct requirements on resolution, latency, temporal modeling, domain coverage, and integration with the language model~\cite{wang2024qwen2,bai2025qwen25vltechnicalreport,wang2022internvideo}. This creates a growing need for SOTA-level visual encoders that can be trained or customized for specific use cases, yet achieving this goal under realistic resource budgets remains challenging.

This difficulty arises from multiple sources. First, the compute cost of visual encoding grows rapidly as visual sequences become longer. High-resolution images, multi-image inputs, and video frames all increase the number of visual tokens, making standard softmax-attention ViTs expensive to train and deploy~\cite{vit,liu2021swin,katharopoulos2020transformers}. Second, strong ViTs often rely on massive web-scale image-text data, yet reproducible and actionable recipes for turning such noisy data into high-quality visual-language supervision remain limited~\cite{schuhmann2022laion,gadre2023datacompsearchgenerationmultimodal,fang2024data,chen2024sharegpt4v}. Simply scaling data can introduce noise, weak alignment, and redundant supervision, making training less efficient. Third, most pretrained ViTs are still trained with fixed-resolution inputs, whereas downstream VLMs often require flexible resolution handling. Existing solutions typically rely on tiling-based processing or additional continual training for dynamic resolution, which introduces extra cost, creates a mismatch between visual pretraining and downstream VLM usage, and may lead to suboptimal performance~\cite{liu2024llavanext,chen2024internvl,dehghani2023patch}.

Together, these barriers make SOTA visual encoders~\cite{tschannen2025siglip2multilingualvisionlanguage,longclip,tulip} inaccessible to much of the broader research and application community. In this report, we revisit ViT design from the perspective of accessibility: can SOTA-level visual encoders be trained, adapted, and deployed under controlled and reproducible resource budgets? Our answer is TuringViT, a VLM-native Vision Transformer designed for efficiency across architecture, data, and training.

\begin{figure}[b]
    \centering
    \includegraphics[width=0.8\textwidth]{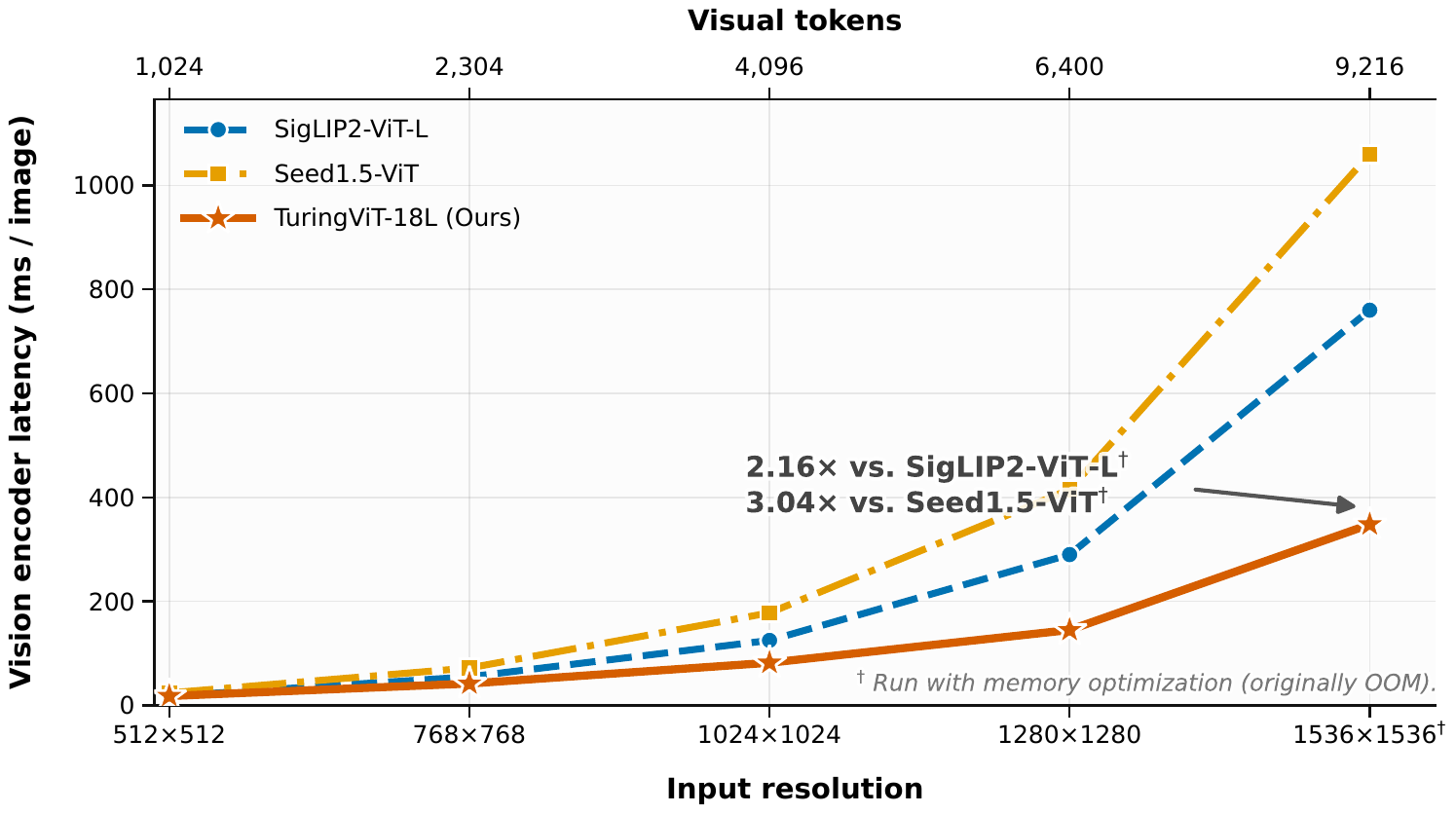}
    \caption{\textbf{Inference latency scaling with input resolution and visual token count.} Latency is measured per image using FP16 TensorRT on an NVIDIA GeForce RTX 3080 Ti with CUDA 11.6; TuringViT shows flatter high-resolution scaling than standard ViT baselines.}
    \label{fig:turingvit_latency_scaling}
\end{figure}

We first address the most fundamental source of cost: the model architecture itself. TuringViT introduces \textbf{Turing Linear Attention (TLA)} as its dominant attention operator, replacing most softmax-attention layers with a linear-complexity alternative tailored to high-resolution and dynamic-resolution visual inputs~\cite{shen2021efficient,choromanski2020rethinking,yang2023gated,wang2020linformer}. TLA combines efficient global context aggregation with sequence-length-aware normalization and an input-dependent output gate, improving stability under variable-length token sequences while preserving local and high-frequency visual responses. To avoid sacrificing explicit token-to-token interaction, TuringViT further retains a small number of standard multi-head softmax-attention layers, inserted periodically among TLA layers. This hybrid design makes linear attention the main computational path while using full attention only sparsely for global routing and token-level interaction. As shown in Figure~\ref{fig:turingvit_latency_scaling}, TuringViT exhibits a much flatter inference-latency curve than standard softmax-based ViTs as the visual sequence length increases, reducing practical sequence-length-to-latency scaling from quadratic to near-linear. This architectural efficiency is crucial for making high-resolution and dynamic-resolution ViT training feasible under controlled resource budgets.

Complementing the efficient architecture, we build \textbf{VISTA-Curation} (\textbf{Vi}sion Data Curation via Image-Video \textbf{S}coring and \textbf{T}emporal \textbf{A}ggregation), a supervision-enhancing multimodal data curation pipeline that enables TuringViT to outperform leading open-source ViT baselines using only 10\% of the data scale. Rather than relying on brute-force data scaling, VISTA-Curation increases the training value of each sample by converting noisy image-language and video-language data into more grounded, discriminative, and temporally informative supervision. For image-language data, it filters low-quality images, generates multiple visually grounded caption candidates, and selects the best caption through relative image-text scoring and textual informativeness ranking. This process suppresses ambiguous or weakly aligned captions while retaining descriptions that are faithful, specific, and useful for CLIP-style alignment. For video-language data, VISTA-Curation decomposes raw videos into compact clips, samples representative frames, filters clips by semantic and motion consistency, and fuses local frame-level details with global temporal semantics. The resulting video captions provide transformation-aware supervision across changes in viewpoint, object state, motion, and scene context. Together, these curated image and video data shift visual-language pretraining from simply using more data to extracting more reliable supervision from each sample. Our scaling-law analysis further shows predictable gains as the curated data scale increases, indicating that TuringViT remains far from saturated.

TuringViT is trained with dynamic resolution throughout the entire visual pretraining process while preserving high training efficiency. Instead of restricting the model to a fixed input size, we allow images with different resolutions and aspect ratios from the beginning. To make this efficient, we adopt a progressive resolution schedule: during the first 80\% of training epochs, we cap the maximum image size at 512, which is already comparable to the maximum resolution supported by many existing ViT encoders; in later stages, we remove this constraint and train with larger images and video data. Benefiting from its linear-complexity architecture, TuringViT maintains efficient training even as visual sequence length increases. This native dynamic-resolution training reduces the need for post-hoc resolution adaptation and leads to consistently stronger downstream VLM performance.

TuringViT delivers strong results across both vision and multimodal evaluations. It achieves an average score of \textbf{83.6} on six zero-shot classification benchmarks and \textbf{78.9} on two retrieval benchmarks, outperforming SigLIP2~\cite{tschannen2025siglip2multilingualvisionlanguage} while using only \textbf{10\%} of the data scale. When integrated into downstream VLMs, TuringViT also brings consistent improvements across multimodal benchmarks. These results show that TuringViT is not merely a faster replacement for standard softmax-based visual encoders, but a stronger, more data-efficient, and more VLM-compatible visual backbone.

%% file: sec/3_methodology.tex
\section{Methodology} \label{sec:methodology}
\begin{figure}[t]
    \centering
    \includegraphics[width=1.0\textwidth]{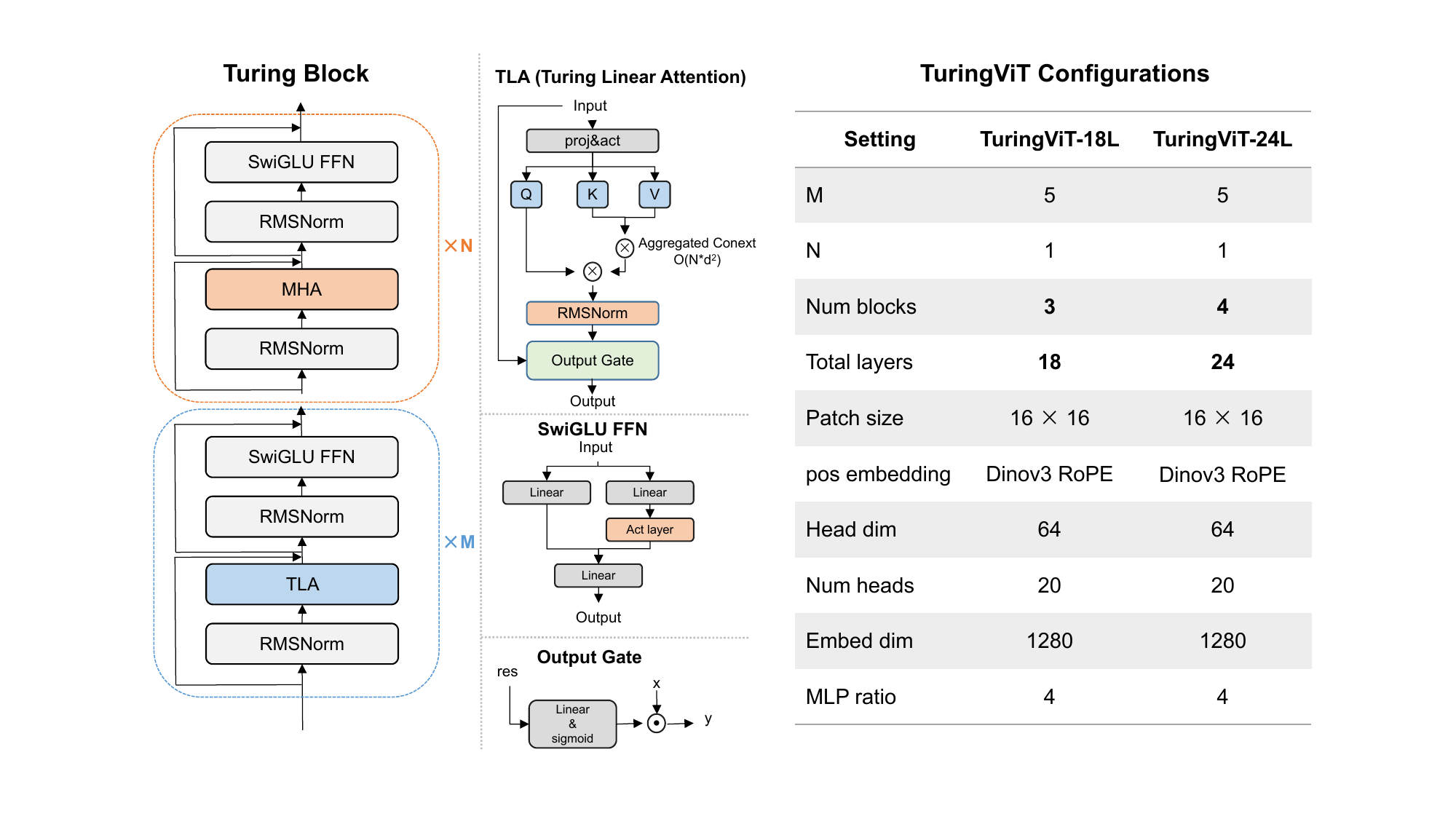}
    \caption{\textbf{TuringViT block architecture and model configurations.} \textbf{(Left)} The backbone is built on repeated Turing Blocks, Turing Linear Attention (TLA) aggregates global context; its sequence-length-aware normalization and input-dependent output gate stabilize variable-token training and preserve high-frequency local details. \textbf{(Right)} Model configurations for TuringViT-18L and TuringViT-24L, which scale in blocks and total layers while sharing identical patch size ($16\times16$), 2D RoPE, embed/head dimension, and MLP ratio.}
    \label{fig:turingvit_structure}
\end{figure}

TuringViT consists of three key components: an efficient visual architecture, a large-scale visual supervision pipeline, and a native dynamic-resolution training paradigm.

First, we introduce \textbf{Turing Linear Attention (TLA)}, a linear-attention-dominant architecture designed for efficient visual sequence modeling as shown in Figure~\ref{fig:turingvit_structure}. Second, we develop \textbf{VISTA-Curation}, a large-scale image-video curation framework that improves supervision quality and data efficiency. Finally, we adopt \textbf{native dynamic-resolution training}, enabling the visual encoder to learn directly from variable-resolution visual inputs throughout pretraining.

Together, these design choices allow TuringViT to achieve strong visual representation quality while remaining efficient and naturally compatible with downstream multimodal systems.
The remainder of this section describes these components in detail.

\subsection{Linear-Attention-Dominant Native-Resolution Architecture}
\label{sec:architecture}

TuringViT is designed to support high-resolution and dynamic-resolution visual inputs for VLM usage. Instead of resizing all images to a fixed square resolution, TuringViT keeps images at native or dynamically selected resolutions and converts them into variable-length visual token sequences with a patch size of $16 \times 16$~\cite{dehghani2023patch}. This design allows the visual encoder to process images with different aspect ratios and resolutions under a unified architecture.

To provide spatial information for variable token grids, we adopt a Dinov3-style~\cite{siméoni2025dinov3} two-dimensional rotary positional embedding~\cite{su2024roformer,heo2024rotary}. The 2D RoPE is applied to the query and key features according to their spatial coordinates on the patch grid. Compared with fixed absolute positional embeddings, this design naturally supports different image sizes and aspect ratios without requiring position-embedding interpolation.

The main computation module of TuringViT is~\textbf{Turing Linear Attention (TLA)}. Given an input token sequence $X \in \mathbb{R}^{N \times C}$, TLA first projects the hidden states into query, key, and value features $Q,K,V$, where $G(X)$ denotes the input-dependent output gate, $\sigma(\cdot)$ is the sigmoid function, $N$ is the input sequence length, and $d$ is the head dimension. We use SiLU as the kernel function $\phi(\cdot)$ and apply it to QKV features. The TLA output is computed as

\begin{equation}
\begin{aligned}
G(X) &= \sigma(XW_g), \\
\mathrm{TLA}(X)
&=
G(X)
\odot
\frac{
\phi(Q)\left(\phi(K)^{\top}\phi(V)\right)
}{
N\sqrt{d}
}.
\end{aligned}
\end{equation}
The normalization by $N\sqrt{d}$ improves numerical stability under dynamic-resolution inputs with variable token lengths. Since linear attention aggregates global context efficiently, it may also smooth local details and high-frequency visual cues. The input-dependent sigmoid output gate provides an additional element-wise modulation path for preserving input-dependent local and high-frequency responses, thereby complementing the low-cost global mixing of linear attention.

Although TLA is used as the dominant attention operator, TuringViT still retains a small number of standard multi-head softmax attention layers. Specifically, we organize the network with a repeated Turing Block, where each block contains five TLA layers followed by one vanilla MHA layer:
\begin{equation}
\mathcal{B}_{\mathrm{Turing}}
=
[\mathrm{TLA},
\mathrm{TLA},
\mathrm{TLA},
\mathrm{TLA},
\mathrm{TLA},
\mathrm{MHA}].
\end{equation}

In this design, vanilla MHA is sparsely inserted at the layer level rather than used in every Transformer layer. The TLA layers provide efficient global context aggregation for long visual sequences, while the periodically inserted MHA layers maintain explicit token-to-token interaction. Therefore, TuringViT is not a strictly linear-attention-only model; its practical efficiency comes from making linear attention the dominant path while using full softmax attention only periodically.

All layers follow a pre-normalization Transformer layout with RMSNorm~\cite{zhang2019root}. Each attention layer is paired with a SwiGLU feed-forward network~\cite{shazeer2020glu}, following the block primitives commonly used in modern large language models. This gives the visual encoder a simple and VLM-friendly structure while keeping the computation efficient for high-resolution inputs.

We instantiate two model variants. \textbf{TuringViT-18L} contains three Turing Blocks, resulting in 18 total layers, including 15 TLA layers and 3 MHA layers. This variant is designed as an efficient and deployment-friendly backbone with strong throughput under dynamic-resolution inputs. \textbf{TuringViT-24L} contains four Turing Blocks, resulting in 24 total layers, including 20 TLA layers and 4 MHA layers. This larger variant further improves representation capacity while keeping the computation dominated by the linear-attention path.

\begin{figure*}[t]
    \centering
    \includegraphics[width=\textwidth]{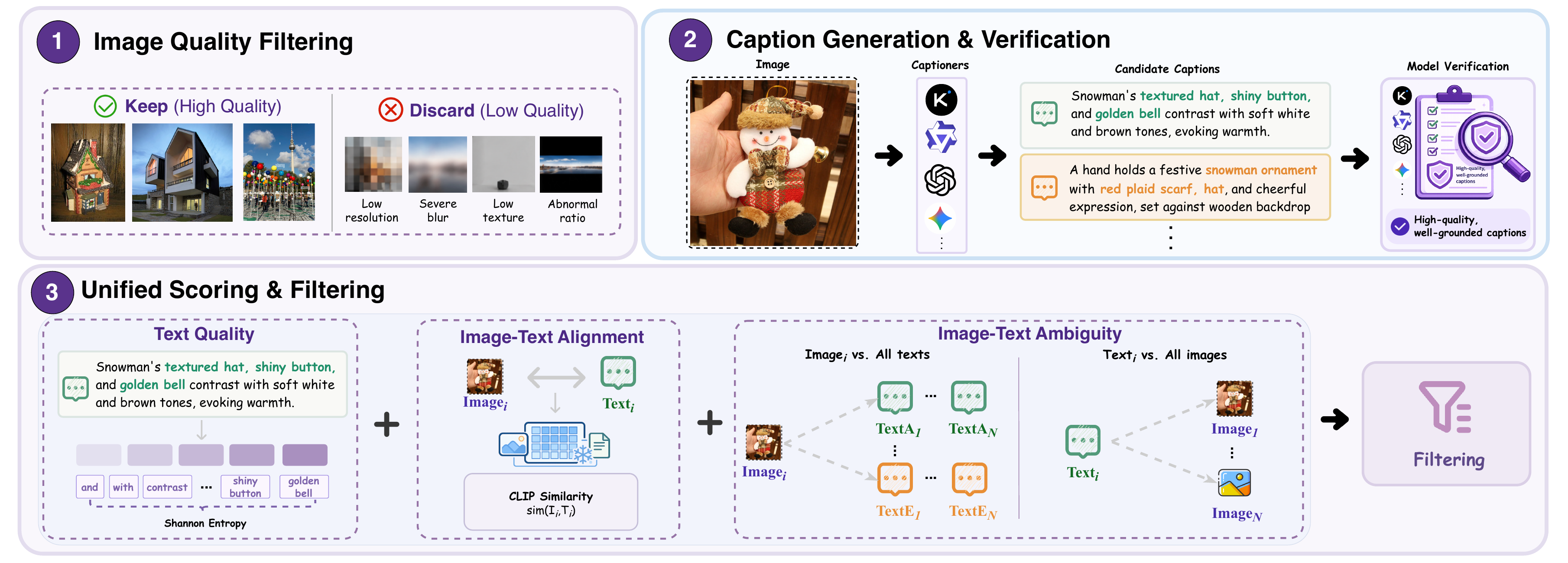}
    \caption{
    \textbf{Overview of the image-language curation engine.} The pipeline filters low-quality images, generates and verifies diverse candidate captions, and ranks all verified candidates in a shared comparison pool. The unified scoring function combines relative visual-language discriminativeness with textual informativeness, selecting captions that are visually grounded, semantically rich, and discriminative for CLIP-style pre-training.
    }
    \label{fig:pipeline}
\end{figure*}
\subsection{Vision Data Curation via Image-Video Scoring and Temporal Aggregation}
\label{sec:data_curation}

With an efficient vision encoder established, we further build \textbf{VISTA-Curation}, a data curation pipeline that selects visually grounded,
semantically rich, and discriminative image-video captions for CLIP-style pre-training.

\subsubsection{Image-Language Data Curation}
As shown in Figure~\ref{fig:pipeline}, image-language data curation aims to convert noisy web image-text pairs into high-quality supervision with stronger visual grounding and higher semantic density. Instead of directly using raw alt-text, we first filter low-quality images, then generate and verify multiple candidate captions to obtain descriptions that are more visually faithful, informative, and suitable for CLIP-style alignment.

\noindent\textbf{Stage 1: Image Quality Filtering.}
We first collect large-scale image sources from public datasets such as DataComp-1B~\cite{gadre2023datacompsearchgenerationmultimodal} and LAION-2B~\cite{schuhmann2022laion}, and perform rule-based filtering to remove low-quality visual samples before caption generation. Images with extremely low resolution, severe blur, low texture, or abnormal ratios are discarded. This step prevents the captioning models from producing unreliable descriptions for images that are visually ambiguous or lack sufficient detail.

\noindent\textbf{Stage 2: Caption Generation and Verification.}
We use carefully designed prompts to generate visually grounded and informative captions while suppressing hallucinations. For each retained image, multiple multimodal models and prompt variants produce diverse candidate captions, which are cross-verified for grounding and factual consistency. This converts weak web annotations into richer, semantically broader supervision.

\noindent\textbf{Stage 3: Unified Scoring and Filtering.}
Given an image $I_i$, the caption generation and verification stages produce a filtered candidate set $\mathcal{C}_{I_i}$ from different captioners and prompt variants. Instead of scoring each candidate independently or only comparing captions within the same image, we collect all candidates that pass the preceding filtering stages into a shared comparison pool $\mathcal{P}$. This allows each image-caption candidate $(I_i, c)$ to be evaluated relative to both its intra-image alternatives and candidates from other images. Such relative scoring strengthens both image-side and text-side filtering: a high-quality caption should not only align well with its paired image, but also be sufficiently discriminative, avoiding ambiguous descriptions that are interchangeable across different images or captions.

Following the relative scoring principle of s-CLIPLoss~\cite{wang2024cliploss}, we evaluate each
image-caption candidate in a shared comparison space, without directly using the
absolute image-text similarity as the only criterion. Given an image $I_i$ and its
filtered candidate caption set $\mathcal{C}_{I_i}$, we score each candidate caption
$c \in \mathcal{C}_{I_i}$ by a relative visual-language quality function:
\begin{equation}
s_{\mathrm{vl}}(I_i, c)
=
-\ell_{\mathrm{sCLIP}}(I_i, c; \mathcal{P}),
\end{equation}
where $\ell_{\mathrm{sCLIP}}(\cdot)$ denotes the relative image-text loss computed over
a sampled comparison pool $\mathcal{P}$. Intuitively, this score considers not only
the alignment between the current image and its paired caption, but also how well the
caption can be distinguished from other captions and how selectively the image matches
its paired text compared with other images.

To further emphasize textual informativeness, we compute an additional text quality
score $s_{\mathrm{text}}(c)$ using caption-level statistics such as Shannon entropy. The final ranking score is
defined as
\begin{equation}
S(I_i,c)
=
s_{\mathrm{vl}}(I_i,c)
+
\alpha \cdot \mathrm{Norm}\big(s_{\mathrm{text}}(c)\big),
\end{equation}
where $\alpha$ balances relative visual-language discriminativeness and textual
informativeness.

Different from scoring only a single generated caption, we apply this ranking process
to all captions that pass the preceding filtering stages. These candidates include
object-centric, attribute-focused, scene-level, relation-aware, and fine-grained
captions. By comparing all filtered candidates within the same relative scoring space,
our pipeline jointly strengthens both image-side and text-side filtering: visually
misaligned captions are suppressed, while generic, redundant, ambiguous, or weakly
discriminative captions are also penalized. For each image $I_i$, we select the
caption with the highest final score:
\begin{equation}
c_i^*
=
\arg\max_{c \in \mathcal{C}_{I_i}} S(I_i,c).
\end{equation}
\begin{figure*}[t]
    \centering
    \includegraphics[width=0.9\textwidth]{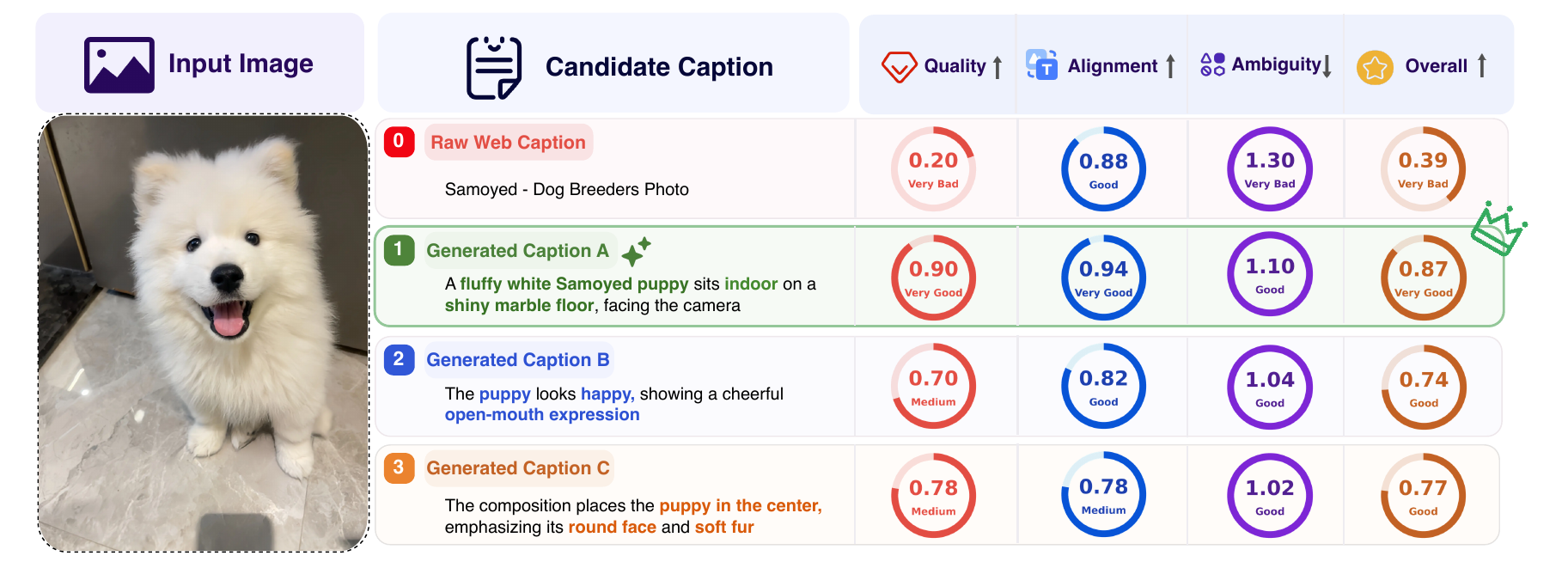}
    \caption{
    \textbf{Qualitative example of multi-candidate recaptioning and scoring.} The selected caption is more detailed, less ambiguous, and better aligned with the image than the original web caption.
    }
    \label{fig:recaption_scoring_example}
\end{figure*}
This strategy encourages the selected captions to be not only visually grounded, but
also semantically specific and discriminative. As shown in
Figure~\ref{fig:recaption_scoring_example}, the raw web caption can obtain a reasonable
alignment score because it is category-related, but its limited textual detail and high
ambiguity lead to a lower overall ranking. In contrast, the selected generated caption
achieves the best overall score by providing richer, visually grounded details while
remaining well aligned with the image and less interchangeable with other samples.

\begin{figure}[H]
    \centering
    \includegraphics[width=1\textwidth]{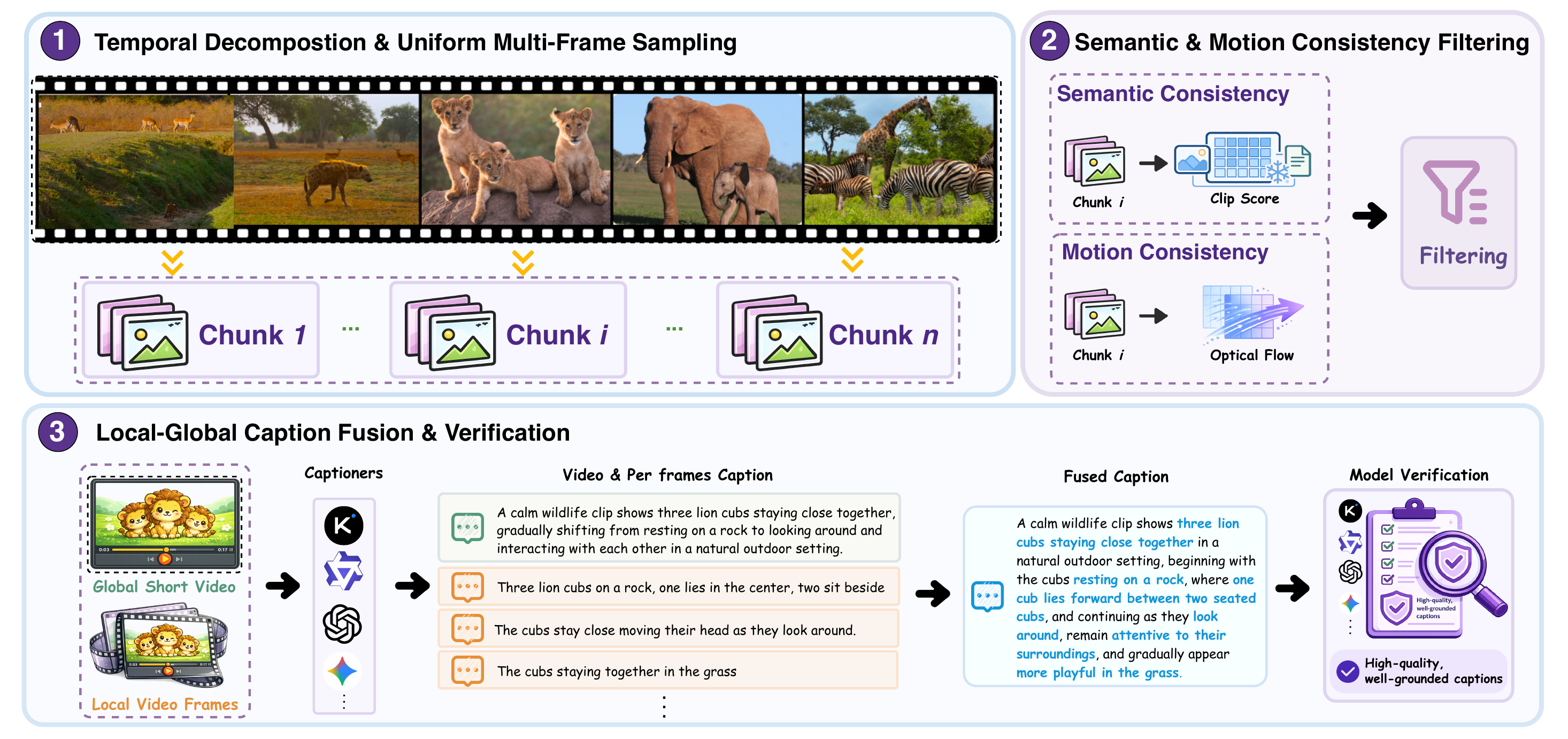}
    \caption{
    \textbf{Overview of the proposed video-language data curation pipeline.}
    Raw videos are decomposed into temporally contiguous short videos, from which uniformly sampled frames are organized as frame chunks. Each chunk is filtered using semantic consistency from pairwise CLIP frame similarities and motion consistency from optical-flow statistics. For the retained short videos, local frame-level captions and global video-level summaries are fused and verified by a VLM to remove low-quality or insufficiently grounded captions, yielding concise and temporally grounded video-language pairs for CLIP-style training.
    }
    \label{fig:video_caption_pipeline}
\end{figure}

\subsubsection{Video-Language Data Curation}
\label{sec:video_data_curation}

Beyond static image-text pairs, videos provide transformation-aware supervision by observing the same semantic entities across time, capturing objects, scenes, and actions under changing viewpoints, scales, poses, occlusions, lighting conditions, and temporal states. Such continuous observations help ViT learn more invariant, object-centric, and context-aware representations beyond isolated snapshots. However, raw videos are highly redundant, weakly annotated, and expensive to use directly, while simply sampling more frames often brings limited additional supervision due to adjacent-frame redundancy. As illustrated in Figure~\ref{fig:video_caption_pipeline}, we therefore curate videos into compact video-language pairs through temporal decomposition, uniform frame sampling, semantic-motion filtering, and local-global caption fusion, transforming long and noisy videos into concise, temporally grounded, and information-rich supervision for CLIP-style video training.

\noindent\textbf{Stage 1: Temporal Decomposition and Uniform Frame Sampling.}
As shown in Figure~\ref{fig:video_caption_pipeline}, each raw video is first decomposed into temporally contiguous short videos based on visual-content changes with the assistance of multimodal models. For each retained short video, we uniformly sample multiple frames at fixed temporal intervals and organize them as frame chunks, following the same protocol used in CLIP-style video training.

\noindent\textbf{Stage 2: Semantic and Motion Consistency Filtering.}
For each chunk, we evaluate the sampled frames from two complementary aspects: semantic consistency and motion consistency. The former is measured by encoding sampled frames with a CLIP image encoder and computing pairwise frame-to-frame similarities, while the latter is measured by computing optical-flow statistics between adjacent sampled frames. Short videos with excessively large motion or near-zero motion are filtered out, and only those with coherent semantics and informative temporal changes are retained.

\noindent\textbf{Stage 3: Local-Global Caption Fusion and Verification.}
For each retained short video, we generate local frame-level captions and a global video-level summary. The local captions describe fine-grained visual details from the sampled frames, while the global summary captures the overall semantics and temporal dynamics of the short video. We then fuse them into a single caption and apply a VLM-based verification step to remove low-quality captions, especially those with insufficient grounding in the video content. The final retained captions are used as reliable video-language pairs for CLIP-style training.

\begin{table*}[htbp]
\centering
\caption{\textbf{Setup for the progressive multi-stage ViT pre-training.} The H800 GPU-hours\textsuperscript{*} indicates total training cost, measured as the number of H800 GPUs multiplied by wall-clock training hours. All results in this table correspond to the TuringViT-24L pre-training setup.}
\label{tab:pretrain_recipe}
\small
\setlength{\tabcolsep}{5pt}
\renewcommand{\arraystretch}{1.15}
\resizebox{\textwidth}{!}{%
\begin{tabular}{lcccc}
\toprule
\textbf{Configuration}
& \makecell{\textbf{Stage 1}}
& \makecell{\textbf{Stage 2}}
& \makecell{\textbf{Stage 3}}
& \makecell{\textbf{Stage 4}} \\
\midrule
Data type
& Unlabeled images
& Image--text pairs
& Image--text pairs
& Image/Video--text pairs \\
Samples
& 55M
& 850M
& 850M
& 2M \\
Resolution
& $256{\times}256$
& Range-constrained
& Native
& Native \\
Objective
& MIM distillation
& SigLIP + SuperClass
& SigLIP + SuperClass
& SigLIP + SuperClass \\
Batch size
& 16,384
& 65,536
& 8,192
& 1,024 \\
Peak LR
& $3.0{\times}10^{-3}$
& $5.0{\times}10^{-4}$
& $5.0{\times}10^{-6}$
& $1.0{\times}10^{-6}$ \\
Final LR
& $3.0{\times}10^{-6}$
& $6.0{\times}10^{-6}$
& $6.0{\times}10^{-8}$
& $2.0{\times}10^{-8}$ \\
H800 GPU-hours\textsuperscript{*} 
& 9984
& 30720
& 12288
& 460 \\
\bottomrule
\end{tabular}%
}
\end{table*}
\vspace{-2.5em}
\subsection{TuringViT Pre-training Stage}
\label{sec:training}

We pre-train TuringViT with a four-stage recipe designed for VLM-native visual encoding. The goal is not to introduce complex objectives, but to progressively expose the visual encoder to the same types of inputs used by downstream VLMs: dynamic-resolution images, high-resolution visual content, and mixed image-video data. Both TuringViT-18L and TuringViT-24L follow this recipe. The detailed training configuration is summarized in Table~\ref{tab:pretrain_recipe}.

\noindent\textbf{Stage 1: MIM-based visual initialization.}
We first pre-train the randomly initialized TuringViT backbone with masked image modeling (MIM) before large-scale image--text alignment~\cite{fang2024eva}. Following the EVA-style feature reconstruction paradigm, we use EVA02-CLIP-E~\cite{sun2023eva} as the teacher model. For each image, 40\% of the image patches are randomly masked, and the student visual encoder is trained to reconstruct the teacher CLIP features at the masked locations. By reconstructing masked patch features, MIM supplies dense local supervision that encourages the linear-attention-dominant backbone to preserve fine-grained geometry and high-frequency visual cues, counteracting linear attention's bias toward global context aggregation.

\noindent\textbf{Stage 2: Range-constrained dynamic-resolution image-text pre-training.}
After visual initialization, we train TuringViT with large-scale image-text contrastive learning. Different from fixed-resolution CLIP pre-training, we preserve the original aspect ratio and resize each image only when necessary, clamping its longer side to the $[256, 512]$ range. This range-constrained dynamic-resolution strategy exposes the model to variable-length visual token sequences from the beginning of contrastive pre-training, while keeping the training cost controlled. Even under this range-constrained dynamic-resolution setting, this stage already involves substantially longer visual token sequences than conventional low-resolution pre-training. Because TuringViT uses linear attention as the dominant computation path, it can efficiently train on these variable-length sequences while maintaining high throughput.
We initialize the visual encoder from Stage 1 and the text encoder from the EVA02-CLIP-E~\cite{sun2023eva} text encoder. For each image-text pair, the visual patch features are aggregated by attention pooling into a single 1024-dimensional image embedding. We align image and text embeddings by jointly optimizing the SigLIP loss~\cite{siglip} and the SuperClass loss~\cite{huang2024classification}.

\noindent\textbf{Stage 3: Native-resolution image-text refinement.}
We further refine TuringViT with image-text contrastive training at native resolution. Unlike the range-constrained setting in Stage 2, this stage preserves the original aspect ratio and resolution of each image whenever possible, making visual preprocessing consistent with downstream VLM usage. The objective remains unchanged, allowing the model to strengthen high-resolution image-text alignment and learn finer visual details while benefiting from the efficiency of its linear-attention-dominant backbone.

\noindent\textbf{Stage 4: Mixed Image/Video-Text Alignment.}
Finally, we adapt TuringViT with mixed image-text and curated video-text training, where video-text pairs are produced by the pipeline in Section~\ref{sec:video_data_curation}. This compact video stage extends the encoder from static image representation to transformation-aware image-video representation, improving transferability for both image and video tasks. For a video with uniformly sampled $T$ frames, we encode each frame with the image encoder and aggregate frame features by attention pooling followed by temporal mean pooling:
\[
v^{\text{vid}} = \text{Norm}\left( \frac{1}{T} \sum_{t=1}^{T} \text{AttnPool}(F_t) \right).
\]

%% file: sec/4_experiments.tex
\section{Experiments}\label{sec:experiments}

We evaluate two TuringViT variants, TuringViT-18L and TuringViT-24L, both trained under the training paradigm described in Section~2.3 with dynamic-resolution image inputs. Our main models are trained on approximately 0.85B image-text pairs. We compare TuringViT with representative vision-language encoders at similar or larger scales, including MobileCLIP2-L~\cite{faghri2025mobileclip2}, SigLIP2-L~\cite{tschannen2025siglip2multilingualvisionlanguage}, and Seed1.5-ViT~\cite{guo2025seed1}. We conduct a comprehensive evaluation covering image zero-shot classification and robustness, image-text retrieval, frozen dense prediction transfer, ablation studies on data strategy, video training, and data scaling, as well as downstream VLM performance under a controlled vision-language training pipeline. These experiments are designed to examine not only the standalone visual representation quality of TuringViT, but also its transferability to dense perception, video-enhanced training, and multimodal reasoning scenarios.

\subsection{Evaluating TuringViT in Zero-shot Classification and Retrieval}
\label{sec:zero_shot_results}

For zero-shot evaluation, we assess TuringViT on both image-level recognition and cross-modal retrieval benchmarks. For zero-shot classification, we evaluate models on ImageNet-1K~\cite{deng2009imagenet}, ImageNet-v2~\cite{recht2019imagenet}, ObjectNet~\cite{barbu2019objectnet}, ImageNet-A~\cite{hendrycks2021natural}, ImageNet-R~\cite{hendrycks2021many}, and ImageNet-Sketch~\cite{wang2019learning}, and use the average score across these six datasets as the main robustness metric. We further evaluate zero-shot image-text retrieval on MS-COCO~\cite{lin2015microsoftcococommonobjects} and Flickr30K~\cite{plummer2016flickr30kentitiescollectingregiontophrase}, covering both standard recognition robustness and image-text alignment.

\begin{table*}[b]
\centering
\small
\setlength{\tabcolsep}{4pt}
\renewcommand{\arraystretch}{1.08}
\caption{
    \textbf{Zero-shot performance across classification and retrieval benchmarks.} All values are percentages and higher is better. Best results are in bold.}
\label{tab:model_comparison}
\begin{tabular*}{\textwidth}{@{\extracolsep{\fill}}lccccc@{}}
\toprule
\textbf{Model}
& \textbf{MobileCLIP2-L}
& \textbf{SigLIP2-L}
& \textbf{Seed1.5-ViT}
& \textbf{TuringViT-18L}
& \textbf{TuringViT-24L} \\
\midrule
\rowcolor{gray!10}\multicolumn{6}{l}{\textbf{Model configuration}} \\
Resolution & 224 & 384 & dyn. & dyn. & dyn. \\
Pretraining data & 1.9B & 10B & 4.8B & 0.85B & 0.85B \\
\midrule
\rowcolor{gray!10}\multicolumn{6}{l}{\textbf{Zero-shot classification}} \\
\textbf{Avg} & 77.0 & 83.4 & 82.5 & 82.7 & \textbf{83.6} \\
ImageNet-1K & 81.9 & 83.1 & 83.6 & 83.1 & \textbf{83.9} \\
ImageNet-v2 & 74.7 & 77.4 & 77.6 & 77.6 & \textbf{78.0} \\
ObjectNet & 75.3 & \textbf{84.4} & 79.2 & 79.7 & 81.1 \\
ImageNet-A & 69.0 & 84.3 & 85.5 & 87.9 & \textbf{89.7} \\
ImageNet-R & 91.7 & \textbf{95.7} & 95.2 & 95.0 & 95.3 \\
ImageNet-Sketch & 69.8 & \textbf{75.5} & 74.1 & 72.8 & 73.6 \\
\midrule
\rowcolor{gray!10}\multicolumn{6}{l}{\textbf{Zero-shot retrieval}} \\
\textbf{Avg} & 72.5 & 76.7 & -- & 78.9 & \textbf{79.4} \\
COCO T$\rightarrow$I & 51.6 & 55.3 & -- & 58.6 & \textbf{59.4} \\
COCO I$\rightarrow$T & 69.0 & 71.4 & -- & 75.9 & \textbf{76.0} \\
Flickr30K T$\rightarrow$I & 77.2 & 85.0 & -- & 85.7 & \textbf{86.0} \\
Flickr30K I$\rightarrow$T & 92.0 & 95.2 & -- & 95.3 & \textbf{96.3} \\
\bottomrule
\end{tabular*}
\vspace{-0.4em}
\end{table*}
Compared with SigLIP2-L~\cite{tschannen2025siglip2multilingualvisionlanguage}, TuringViT-24L achieves a higher average zero-shot classification score while using only about 10\% of the pretraining data. The gains are particularly clear on ImageNet-1K, ImageNet-v2, and ImageNet-A, indicating that dynamic-resolution training and the proposed data-centric pretraining recipe lead to strong zero-shot recognition and robustness under challenging visual distribution shifts. We also observe that TuringViT-24L still lags behind SigLIP2-L~\cite{tschannen2025siglip2multilingualvisionlanguage} on ObjectNet, ImageNet-R, and ImageNet-Sketch. These benchmarks emphasize object viewpoint changes, rendition-style distribution shifts, and sketch-like visual abstraction, which are closely tied to the coverage of difficult and out-of-distribution visual samples in pretraining data. Since TuringViT-24L is trained with a much smaller data budget, the amount of such hard visual data is relatively limited. Nevertheless, TuringViT-24L remains competitive on these benchmarks compared with other baselines, and we expect that further scaling the curated pretraining corpus with more viewpoint-diverse, rendition-style, and sketch-like samples can further improve performance. On retrieval benchmarks, our model demonstrates a clear advantage, which we attribute to the construction of accurate and highly aligned image-text pairs in our data pipeline.

\subsection{Evaluating TuringViT as a Frozen Dense Visual Encoder}
\label{sec:frozen_dense_eval}

We further evaluate the transferability of TuringViT to dense prediction tasks, as summarized in Table~\ref{tab:frozen_dense_eval}. To this end, we freeze the pretrained visual encoder and use only the final-layer patch features for downstream dense evaluations. Each model is evaluated at its training resolution; for TuringViT, we use 1024 as the representative resolution, corresponding to the median resolution observed during dynamic-resolution training. We consider three representative dense tasks: monocular depth estimation on NYUv2~\cite{silberman2012indoor}, semantic segmentation on ADE20K~\cite{zhou2017scene}, and zero-shot object tracking on DAVIS-2017~\cite{pont20172017}.

\noindent{\bf Depth estimation.}
For depth estimation, we evaluate frozen TuringViT features on. We attach a lightweight DPT-style~\cite{ranftl2021vision} depth head on top of the frozen visual features. Since NYUv2 is relatively small and large pretrained encoders are prone to overfitting under dense probing, we train the depth head for 20 epochs and report the best validation result.

\noindent{\bf Semantic segmentation.}
For semantic segmentation, we evaluate frozen TuringViT features on ADE20K. We train a lightweight segmentation head on top of the frozen patch features. The head follows a UPerNet-style~\cite{xiao2018unified} design to aggregate multi-scale contextual information and produce per-pixel semantic predictions.

\noindent{\bf Zero-shot tracking.}
For object tracking, we adopt a training-free zero-shot label propagation protocol on DAVIS-2017~\cite{pont20172017}. Instead of training a task-specific head, we use the first-frame object mask as the reference and extract normalized patch features from each video frame with the frozen TuringViT encoder. For each patch in the current frame, we search for its nearest neighbors among the patch features from the most recent 7 frames and propagate the corresponding object labels to the current frame. We report the standard DAVIS $\mathcal{J}\&\mathcal{F}$ score as the evaluation metric.
\vspace{-0.4em}
\begin{table}[H]
\centering
\small
\setlength{\tabcolsep}{6pt}
\renewcommand{\arraystretch}{1.05}
\caption{
\textbf{Evaluation of frozen dense visual representations.}
We freeze each visual encoder and evaluate its transferability on monocular depth estimation, semantic segmentation, and zero-shot video object tracking.
$\downarrow$ indicates lower is better, while $\uparrow$ indicates higher is better.
Best results are in bold.
}
\label{tab:frozen_dense_eval}
\begin{tabular}{@{}lcccc@{}}
\toprule
\textbf{Model}
& \textbf{Input Res.}
& \textbf{NYUv2 Depth}
& \textbf{ADE20K Seg.}
& \textbf{DAVIS-2017 Track.} \\
&
& \textbf{RMSE $\downarrow$}
& \textbf{mIoU $\uparrow$}
& \textbf{$\mathcal{J}\&\mathcal{F}$ $\uparrow$} \\
\midrule
MobileCLIP2-L
& 224
& 53.3
& 37.9
& 25.6 \\
SigLIP2-L
& 384
& 51.6
& 42.3
& 32.4 \\
TuringViT-18L
& 1024
& 53.7
& 43.2
& 47.5 \\
\rowcolor{gray!10}\textbf{TuringViT-24L}
& \textbf{1024}
& \textbf{50.7}
& \textbf{45.0}
& \textbf{48.5} \\
\bottomrule
\end{tabular}
\vspace{-0.8em}
\end{table}

\subsection{Ablation Study}
We conduct extensive ablation studies to verify the effectiveness of the key components in TuringViT-18L. Specifically, we analyze three aspects of our training recipe: the contribution of each data strategy component, including recaptioning, dynamic-resolution training, data augmentation, and data filtering; the effect of introducing video data and image replay in later-stage training; and the scaling behavior with respect to the amount of image-text pretraining data.
\begin{table}[!tbp]
\centering
\small
\setlength{\tabcolsep}{4pt}
\scriptsize
\setlength{\tabcolsep}{3.0pt}
\renewcommand{\arraystretch}{1.12}
\caption{
\textbf{Ablation study of the robust pretraining recipe for TuringViT-18L.}
The average score is computed over ImageNet-1K, ObjectNet, ImageNet-v2, ImageNet-A, ImageNet-R, and ImageNet-Sketch.
}
\label{tab:turingvit_robust_pretraining_ablation}
\resizebox{\linewidth}{!}{%
\begin{tabular}{@{}lcccccccccccc@{}}
\toprule
& \multicolumn{4}{c}{\textbf{Strategy component}} & \multicolumn{1}{c}{\textbf{Scale}} & \multicolumn{7}{c}{\textbf{Zero-shot robustness}} \\
\cmidrule(lr){2-5}\cmidrule(lr){6-6}\cmidrule(l){7-13}
\textbf{Variant} & \textbf{Recap.} & \textbf{Dyn.} & \textbf{Aug.} & \textbf{Filt.} & \textbf{Data}
& \textbf{Avg} & \textbf{IN-1K} & \textbf{ObjNet} & \textbf{IN-v2} & \textbf{IN-A} & \textbf{IN-R} & \textbf{IN-S} \\
\midrule
Base
& $\textcolor{red}{\times}$ & $\textcolor{red}{\times}$ & $\textcolor{red}{\times}$ & $\textcolor{red}{\times}$
& 25M & 49.4 & 56.6 & 36.6 & 49.2 & 45.6 & 66.8 & 41.8 \\
+ Recaption
& $\textcolor{ForestGreen}{\checkmark}$ & $\textcolor{red}{\times}$ & $\textcolor{red}{\times}$ & $\textcolor{red}{\times}$
& 25M & 60.7 & 64.8 & 53.7 & 57.5 & 57.2 & 78.3 & 52.6 \\
+ Dynamic res.
& $\textcolor{ForestGreen}{\checkmark}$ & $\textcolor{ForestGreen}{\checkmark}$ & $\textcolor{red}{\times}$ & $\textcolor{red}{\times}$
& 25M & 73.5 & 77.3 & 62.3 & 69.9 & 75.0 & 90.8 & 65.7 \\
+ Augmentation
& $\textcolor{ForestGreen}{\checkmark}$ & $\textcolor{ForestGreen}{\checkmark}$ & $\textcolor{ForestGreen}{\checkmark}$ & $\textcolor{red}{\times}$
& 25M & 74.4 & 78.7 & 63.1 & 71.1 & 75.4 & 91.2 & 66.8 \\
\rowcolor{gray!10}\textbf{Full data strategy}
& $\textcolor{ForestGreen}{\checkmark}$ & $\textcolor{ForestGreen}{\checkmark}$ & $\textcolor{ForestGreen}{\checkmark}$ & $\textcolor{ForestGreen}{\checkmark}$
& 20M & \textbf{76.8} & \textbf{80.1} & \textbf{68.5} & \textbf{73.2} & \textbf{78.5} & \textbf{91.8} & \textbf{68.6} \\
\bottomrule
\end{tabular}%
}
\vspace{-0.4em}
\end{table}

\noindent{\bf Data strategy:} 
We conduct a cumulative ablation study to quantify the contribution of each component in the proposed robust pretraining recipe.
All variants use TuringViT-18L and are trained for 10 epochs, while progressively enabling recaptioning, dynamic-resolution training, data augmentation, and data filtering.
The results are summarized in Table~\ref{tab:turingvit_robust_pretraining_ablation}.
Starting from the 25M baseline, replacing noisy web captions with recaptioned annotations improves the average robustness score from 49.42 to 60.68, giving an absolute gain of +11.26\%.
This gain is broad across benchmarks and is especially large on ObjectNet (+17.09\%), indicating that higher-quality text supervision helps the model learn more object-centric and semantically aligned visual representations.
Adding dynamic-resolution training further increases the average score to 73.49 (+12.81\%), which is the largest single-step improvement in the ablation.
The improvement is particularly strong on ImageNet-A (+17.81\%), suggesting that resolution diversity substantially improves robustness to unusual object appearances, scales, and visual distributions.
Data augmentation provides a smaller but consistent additional gain, increasing the average score from 73.49 to 74.40.
This shows that stronger visual perturbations remain useful even after improving caption quality and resolution diversity.
Finally, data filtering reduces the training set from 25M to 20M image-text pairs but further improves the average score to 76.78 (+2.38\%).
The largest gain from filtering again appears on ObjectNet (+5.34\%), highlighting that removing noisy or weakly aligned samples is especially important for robustness-oriented transfer.

\begin{table*}[!tpb]
\centering
\fontsize{8.0pt}{9.4pt}\selectfont
\setlength{\tabcolsep}{5.2pt}
\renewcommand{\arraystretch}{1.10}
\caption{\textbf{Effect of video data and image replay in later-stage training.} All models use the same image training recipe; higher is better. Best results are in bold.}
\label{tab:video_scaling}
\begin{tabular}{lccccccccc}
\toprule
\textbf{Training strategy}
& \multicolumn{6}{c}{\textbf{Image zero-shot classification}}
& \multicolumn{3}{c}{\textbf{Video classification / retrieval}} \\
\cmidrule(lr){2-7}\cmidrule(l){8-10}
& \textbf{IN-1K} 
& \textbf{Obj.} 
& \textbf{IN-v2} 
& \textbf{IN-A} 
& \textbf{IN-R} 
& \textbf{IN-S}
& \textbf{K400} 
& \textbf{\shortstack{MSR-VTT\\T$\to$V}} 
& \textbf{\shortstack{MSR-VTT\\V$\to$T}} \\
\midrule
Stage 3
& \textbf{83.2} & 75.7 & 77.3 & 84.4 & 94.6 & 72.5
& 64.3 & 43.0 & 39.3 \\
Stage 4 (video only)
& 83.0 & 77.6 & 77.4 & 87.0 & 94.9 & 72.6
& 65.6 & 43.9 & 41.0 \\
\rowcolor{gray!20}\textbf{Stage 4}
& 83.1 & \textbf{79.7} & \textbf{77.6} & \textbf{87.9} & \textbf{95.0} & \textbf{72.8}
& \textbf{66.3} & \textbf{44.5} & \textbf{41.9} \\
\bottomrule
\end{tabular}
\vspace{-0.4em}
\end{table*}

\noindent{\bf Training with Video:}
Table~\ref{tab:video_scaling} compares different Stage 4 training strategies. The first row reports the Stage 3 checkpoint before any additional video-enhanced training, which serves as the baseline. Starting from this checkpoint, using video-only data in Stage 4 improves performance on both image and video benchmarks. For image zero-shot classification, it brings clear gains on robustness-oriented benchmarks such as ObjectNet (+1.9) and ImageNet-A (+2.6), suggesting that video data provides complementary supervision through viewpoint changes, motion patterns, object transformations, and more diverse real-world contexts. Meanwhile, video-only training also improves video classification and retrieval, increasing K400~\cite{kay2017kinetics} from 64.3 to 65.6, MSR-VTT~\cite{xu2016msr} text-to-video retrieval from 43.0 to 43.9, and MSR-VTT video-to-text retrieval from 39.3 to 41.0. However, video-only Stage 4 training is not optimal: it slightly decreases ImageNet-1K performance and remains below the mixed image-video strategy on most benchmarks. When Stage 4 combines video data with image replay, the model achieves the best overall performance. It further improves ObjectNet to 79.7 and ImageNet-A to 87.9, while also achieving the best video results, with 66.3 on K400, 44.5 on MSR-VTT text-to-video retrieval, and 41.9 on MSR-VTT video-to-text retrieval. These results indicate that video data improves transformation-aware and out-of-distribution robustness, while mixing image data helps preserve the broad static-image alignment learned during large-scale image-text pretraining.

\begin{figure*}[t]
    \centering
    \includegraphics[width=0.8\textwidth]{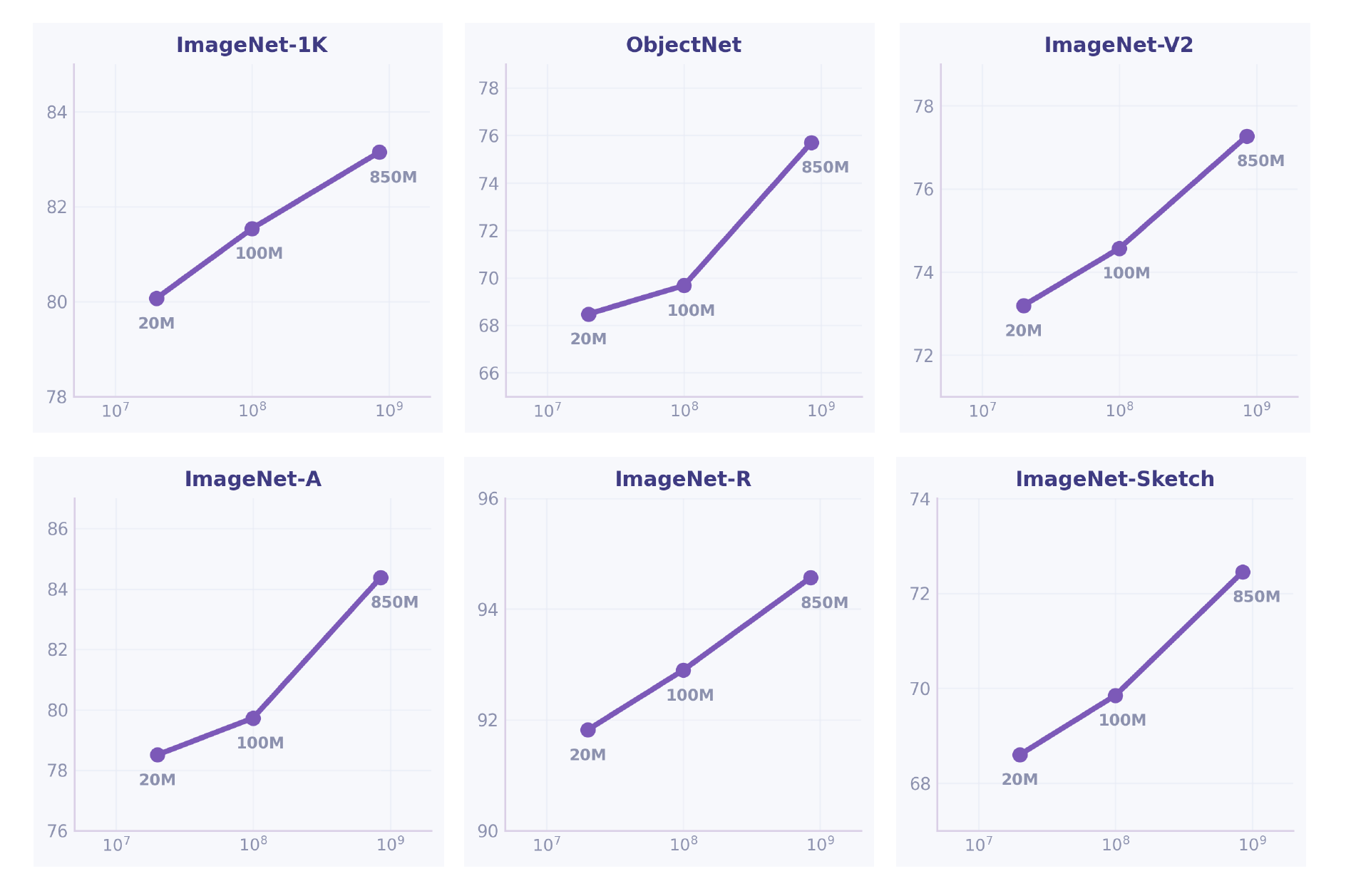}
    \caption{
        \textbf{Training data scaling results of TuringViT-18L.}
        We report zero-shot performance as the amount of pretraining data increases from 20M to 850M image-text pairs. 
        All models are trained for 10 epochs using the same recipe, including recaptioning, dynamic-resolution training, data augmentation, and data filtering.
    }
    \label{fig:data_scaling}
\end{figure*}


\noindent{\bf Data scaling:} 
We conduct a controlled data-scaling study with TuringViT-18L under the same training recipe, including recaptioning, dynamic-resolution training, data augmentation, and filtering. All models are trained for 10 epochs on 20M, 100M, and 850M image-text pairs, respectively, and evaluated on six zero-shot classification benchmarks: ImageNet-1K, ObjectNet, ImageNet-v2, ImageNet-A, ImageNet-R, and ImageNet-Sketch. As shown in Figure~\ref{fig:data_scaling}, performance improves consistently as the amount of training data increases.The gains are especially pronounced on robustness-oriented benchmarks such as ObjectNet and ImageNet-A, indicating that data scaling primarily improves the model's ability to handle distribution shifts. The curve shows no clear saturation up to 850M samples, suggesting that further scaling of curated training data may continue to bring additional improvements.
This ablation demonstrates that for a CLIP-like model, increasing the number of training examples from 20M to 850M yields significant and systematic improvements across a wide range of robustness benchmarks. The gains follow a power law, and no saturation is observed at 850M, suggesting that further scaling of data (and model capacity) would continue to improve performance.

\subsection{Evaluating TuringViT in Downstream VLMs}

\begin{table*}[!t]
\centering
\small
\setlength{\tabcolsep}{6pt}
\renewcommand{\arraystretch}{1.12}
\caption{\textbf{Comparison of downstream VLM performance using different visual encoders under the same VLM training pipeline.}  Best results are marked in bold.}
\label{tab:stem_puzzle_instruct}
\begin{tabular}{lccc}
\toprule
\textbf{Benchmark}
& \textbf{TuringViT-24L}
& \textbf{SigLIP2-L}
& \textbf{MobileCLIP2-L} \\
\midrule
\rowcolor{gray!10}\multicolumn{4}{l}{\textbf{Grounding \& Counting}} \\
RefCOCO avg~\cite{yu2016modelingcontext,mao2016generationcomprehension}       & \textbf{90.4} & 88.9 & 88.2 \\
CountBench~\cite{paiss2023teaching}        & \textbf{81.1} & 79.9 & 78.6 \\
BLINK~\cite{fu2024blink}             & 48.8 & \textbf{49.1} & 47.3 \\
\midrule
\rowcolor{gray!10}\multicolumn{4}{l}{\textbf{Text-rich Perception}} \\
OCRBench V1~\cite{liu2023ocrbench}       & \textbf{798} & 781 & 776 \\
OCRBench V2 en~\cite{fu2024ocrbenchv2}    & \textbf{40.1} & 36.2 & 35.4 \\
OCRBench V2 cn~\cite{fu2024ocrbenchv2}    & \textbf{32.6} & 32.0 & 31.0 \\
ChartQA test~\cite{masry2022chartqa}      & 79.4 & 79.6 & \textbf{80.6} \\
TextVQA val~\cite{singh2019textvqa}       & \textbf{78.7} & 77.1 & 76.2 \\
DocVQA val~\cite{mathew2021docvqa}        & \textbf{88.5} & 87.3 & 87.8 \\
AI2D test~\cite{kembhavi2016diagram}         & \textbf{77.1} & 74.2 & 74.6 \\
\midrule
\rowcolor{gray!10}\multicolumn{4}{l}{\textbf{Multimodal Reasoning}} \\
MMMU val~\cite{yue2024mmmu}          & \textbf{44.2} & 41.8 & 41.6 \\
MathVision mini~\cite{wang2024mathvision}   & \textbf{28.6} & 24.3 & 24.7 \\
MathVista mini~\cite{lu2023mathvista}    & 60.0 & 62.0 & \textbf{62.2} \\
DynaMath~\cite{zou2024dynamath}          & 28.6 & 27.5 & \textbf{28.7} \\
LogicVista~\cite{xiao2024logicvista}        & \textbf{34.5} & 30.2 & 30.9 \\
\midrule
\rowcolor{gray!10}\multicolumn{4}{l}{\textbf{General Visual Question Answering}} \\
MMBench-EN~\cite{liu2023mmbench}        & 76.3 & \textbf{76.9} & 73.1 \\
MMBench-CN~\cite{liu2023mmbench}        & \textbf{74.4} & \textbf{74.4} & 71.0 \\
RealWorldQA~\cite{xai2024grok15v}       & \textbf{64.7} & 63.7 & 62.5 \\
MMStar~\cite{chen2024mmstar}            & \textbf{57.3} & 55.8 & 56.2 \\
MMVet~\cite{yu2023mmvet}             & \textbf{49.1} & 47.0 & 43.4 \\
HallusionBench~\cite{guan2023hallusionbench}    & \textbf{57.6} & 57.4 & 54.8 \\
MME-RealWorld-Lite~\cite{zhang2024mmerealworld} & \textbf{40.7} & 40.3 & 36.5 \\
SEEDBench~\cite{li2023seedbench}         & \textbf{73.8} & 73.5 & 72.6 \\
\midrule
\rowcolor{gray!10}\multicolumn{4}{l}{\textbf{Video Understanding}} \\
MVBench~\cite{li2023mvbench}           & \textbf{63.8} & 63.1 & 63.7 \\
MLVU~\cite{zhou2024mlvu}              & \textbf{57.6} & 55.6 & 53.5 \\
LongVideoBench~\cite{wu2024longvideobench}    & 50.9 & 51.2 & \textbf{51.4} \\
\bottomrule
\end{tabular}
\vspace{-0.4em}
\end{table*}

To further evaluate whether the advantages of TuringViT transfer to multimodal scenarios, we integrate it into downstream VLMs and compare it with representative open-source visual encoders, including SigLIP2~\cite{tschannen2025siglip2multilingualvisionlanguage} and MobileCLIP2~\cite{faghri2025mobileclip2}. Our goal is not to introduce a new VLM recipe, but to provide a controlled comparison of different ViT backbones under the same vision-language training pipeline. Therefore, all models adopt a standard ViT-MLP-LLM architecture, where the visual encoder is connected to the language model through an MLP projector. Most architectural details follow Qwen2.5-VL~\cite{bai2025qwen25vltechnicalreport}, such as token merger in projector and M-RoPE. For the language model, we use Qwen2.5-1.5B-Instruct~\cite{qwen252024} as the LLM backbone across all experiments.

Since this study focuses on evaluating the visual encoder, we only conduct supervised fine-tuning rather than full-scale VLM pretraining. The VLM training is divided into two stages. In Stage 1, we train the visual encoder and the projector while keeping the LLM frozen, using mainly captioning, OCR, knowledge, and VQA-style data to align visual features with the language model. In Stage 2, we unfreeze all parameters and further train the full model with a broader mixture of instruction data, including STEM, grounding, video, text-only, and GUI-related samples. All training data are collected from publicly available sources. For parts of the data with low quality or weak alignment, we apply data filtering and recaptioning to improve supervision quality.

For a fair comparison, we adapt SigLIP2~\cite{tschannen2025siglip2multilingualvisionlanguage} and MobileCLIP2~\cite{faghri2025mobileclip2} to the same dynamic-resolution VLM setting. Since these models are not originally designed with native dynamic-resolution support, we add 2D RoPE to their visual encoders and allow their visual backbones to be updated during Stage 1. This adaptation helps align their visual features with the LLM under the same dynamic-resolution input format, reducing the possibility that their performance is limited by implementation mismatch rather than backbone quality.

We evaluate the resulting VLMs on a diverse set of multimodal benchmarks that are sensitive to visual encoder quality, including grounding and counting, text-rich perception, multimodal reasoning, general VQA, and video understanding. These benchmarks cover spatial localization, OCR and document perception, diagram and mathematical reasoning, real-world visual question answering, hallucination-related evaluation, and temporal understanding. This suite allows us to examine whether the visual advantages of TuringViT transfer to downstream multimodal tasks.

As shown in Table~\ref{tab:stem_puzzle_instruct}, TuringViT achieves the best performance on most benchmarks under the same VLM setup, with particularly clear advantages in grounding and counting, text-rich perception, and general VQA. These categories are closely tied to the quality of the visual encoder: grounding and counting require accurate object-level and spatial representations, text-rich perception depends on fine-grained high-resolution details, and general VQA benefits from robust visual features across diverse real-world inputs. The strong results in these categories suggest that the benefits of TuringViT's architecture, curated supervision, and native dynamic-resolution training transfer effectively from visual pretraining to downstream VLM usage.

The advantage is less uniform on multimodal reasoning and video understanding. This is expected, as these tasks are not determined by the visual encoder alone. Multimodal reasoning benchmarks often require substantial language-side reasoning, symbolic manipulation, and instruction-following ability after the relevant visual evidence is extracted, making their final scores strongly dependent on the LLM and SFT data mixture. Similarly, long-form video benchmarks place additional demands on temporal aggregation, memory, and sequence-level reasoning, while ViT pretraining is still primarily based on image and short-clip supervision. Nevertheless, TuringViT remains competitive in these settings and achieves leading results on several reasoning and video benchmarks. Overall, these results show that TuringViT is not merely an efficient replacement for softmax-based visual encoders, but a stronger VLM backbone whose gains are most pronounced on tasks where visual representation quality is the primary bottleneck.

%% file: sec/5_application.tex
\section{Applications and Broader Impact}
\label{sec:application}

\subsection{Unified Visual Foundation for VLA 2.0, Intelligent Cockpit, and Iron Robotics}

TuringViT has been integrated into multiple XPeng AI systems, including VLA 2.0 for autonomous driving, multimodal systems for intelligent cockpit and driving-parking integration, and the foundation model of the XPeng Iron robot. These systems operate in different environments and require different forms of visual understanding, but all rely on a strong and efficient visual encoder as their perception foundation.

In \textbf{VLA 2.0}, the visual encoder needs to process multi-camera, multi-frame, and dynamic road-scene observations. It provides visual tokens that support spatial understanding, object-state estimation, and downstream policy decisions. Since the VLA system directly serves real-time vehicle-side decision making, the visual encoder must preserve strong perception capability while satisfying strict inference-efficiency requirements.

In \textbf{intelligent cockpit and driving-parking integrated systems}, the visual encoder supports multimodal understanding of both in-cabin and out-of-cabin environments. These tasks include scene recognition, OCR, navigation-related reasoning, parking and driving scene understanding, and interaction with user instructions and vehicle-control systems. In this setting, the visual representation must not only capture visual content accurately, but also align effectively with language, navigation, and control signals.

In the \textbf{XPeng Iron robot}, the visual encoder further provides the perceptual foundation for embodied intelligence. The robot needs to understand object categories, spatial relations, manipulation-relevant regions, occlusions, and dynamic changes in the environment, and then connect these visual signals with language instructions and action planning. Compared with vehicle-side scenarios, robotic tasks place stronger emphasis on close-range object interaction and action-relevant visual details, requiring fine-grained and transferable visual representations.

Across these systems, TuringViT serves as a shared visual foundation while allowing task-specific variants when needed. Different downstream systems can adjust model size, input resolution, deployment operators, or training data based on the same architecture and training methodology, without redesigning the visual encoder from scratch. This flexibility also extends to deeper VLM integration. Since TuringViT is designed to be VLM-native from the beginning, visual-side modules required by the final VLM can be introduced during ViT training, and a later generation-oriented alignment stage can be added to better match next-token prediction and instruction-following objectives. In XPeng's Turing VLM, such downstream-aware variants reduce integration mismatch, enter multimodal training more smoothly, and achieve further improvements on downstream VLM benchmarks.

\subsection{Efficient Scaling and Broader Impact}

The practical significance of TuringViT is not limited to any single downstream system. It is efficient to deploy within XPeng's Turing chip ecosystem, but its broader value is that it provides a scalable path for efficient ViT development. As VLM and VLA systems move toward higher-resolution perception, multi-image inputs, and video understanding, the length of visual sequences will continue to increase. If visual encoders continue to rely primarily on standard softmax attention, such scaling will quickly introduce substantial training and inference cost. By using a more efficient form of visual sequence modeling, TuringViT makes the continued scaling of high-resolution and dynamic-resolution ViTs more practical.

This scaling capability appears not only in the model architecture, but also in the data and training methodology. Our scaling-law analysis shows that TuringViT achieves clear and predictable gains as the curated data scale increases, indicating that the model is still far from saturated. This suggests that TuringViT is not merely a specific model instance that works at the current data and model scale, but a framework that can continue to improve as data, model capacity, and task complexity grow.

More importantly, the core methodology of TuringViT does not depend on XPeng's internal hardware or private systems. Although TuringViT is friendly to deployment on Turing chips, its efficient visual architecture, data construction pipeline, and dynamic-resolution training strategy are general and transferable to other hardware platforms and application scenarios. For research and application teams across the industry, this means that they do not have to rely solely on a small number of large-scale pretrained general-purpose visual encoders. They can instead follow a similar methodology to train, customize, and deploy strong visual foundation models under controlled resource budgets.

Therefore, the application value of TuringViT is twofold. Internally, it supports XPeng's autonomous driving, cockpit, and robotics systems as a unified visual foundation. Externally, it provides the broader multimodal AI community with a reproducible, scalable, and deployable path for visual encoder development. \textbf{TuringViT turns SOTA-level ViTs from a capability accessible only to high-resource teams into a foundation module that more teams can build, customize, and benefit from}.

%% file: sec/6_conclusion.tex
\section{Conclusion and Future Work}
\label{sec:conclusion_future_work}

This technical report presents \textbf{TuringViT}, a VLM-native visual foundation model designed to make strong customized Vision Transformers more accessible under practical training and deployment constraints. Instead of relying on fully quadratic softmax attention and fixed-resolution pretraining, TuringViT combines a linear-attention-dominant backbone, native dynamic-resolution training, and a supervision-rich image-video curation pipeline. This design enables efficient high-resolution visual encoding while preserving strong representation quality for downstream VLM and VLA applications.

Our experiments show that TuringViT achieves competitive or superior performance against strong open-source ViT baselines while using substantially less training data. The model also demonstrates favorable latency scaling under high-resolution and dynamic-resolution inputs, making it suitable for real-world multimodal systems where both visual fidelity and efficiency are critical. Beyond benchmark performance, the results suggest that careful architecture design and data curation can significantly improve the data efficiency of visual foundation model training.

We believe TuringViT provides a practical step toward democratizing the training and customization of SOTA-level visual encoders. In future work, we plan to further scale the curated image-video data engine, strengthen temporal and embodied visual understanding, and explore broader integration with VLM/VLA systems in autonomous driving, robotics, and intelligent interaction scenarios. We hope this report encourages continued research on efficient, VLM-native visual foundation models and helps visual encoders play a larger role in next-generation multimodal applications.

%% file: sec/7_contributors.tex
\section{Contributors}
We sincerely thank every member of the team for their dedication and valuable contributions. This work reflects our ongoing efforts in advancing VLM/VLA-related applications, and we hope that TuringViT will play an increasingly important role in this direction. In addition, our VISTA data curation pipeline is designed to lower the barrier to training visual foundation models.

\textbf{Advisors}: Hang Zhang, Xianming Liu

\textbf{Project Lead}: Qiman Wu

\textbf{Contributors}: 
Hanlin Chen\textsuperscript{*}
, Lyujie Chen\textsuperscript{*}
, Rui Xin\textsuperscript{*}
, Jianlei Zheng
, Mingyuan Wang
, Jiahui Hu
, Da Zhu
, Yuecheng Ma
, Yuhua Wei
, Yizhao Wang
, Hua Zhou
, Yuheng Zhang
, Anhua Liu
, Shaman Tang
, Yue He
, Pengfei Diao
, Shuang Su
, Haotong Xin\textsuperscript{\dag}
, Weichao Huang

\textsuperscript{*}Core contribution. The first three authors are listed in alphabetical order.

\textsuperscript{\dag}Research Intern at XPENG.